\documentclass[a4paper]{article}
\usepackage{graphicx}
\usepackage{INTERSPEECH2020}

\title{End-to-End Code Switching Language Models for Automatic Speech Recognition}
\name{Ahan M R $^1$, Shreyas Sunil Kulkarni$^2$}
\address{
  $^1$ Dept. of Mathematics, BITS Pilani, K K Birla Goa Campus\\
  $^2$ Dept. of Computer Science, BITS Pilani, Hyderabad Campus}
\email{f20160487@goa.bits-pilani.ac.in, f20160649@hyderabad.bits-pilani.ac.in}

\begin{document}

\maketitle
\begin{abstract}
  
  In this paper, we particularly work on the code-switched text, one of the most common occurrences in the bilingual communities across the world. Due to the discrepancies in the extraction of code-switched text from an Automated Speech Recognition(ASR) module, and thereby extracting the monolingual text from the code-switched text, we propose an approach for extracting monolingual text using Deep Bi-directional Language Models(LM) such as BERT and other Machine Translation models, and also explore different ways of extracting code-switched text from the ASR model. We also explain the robustness of the model by comparing the results of Perplexity and other different metrics like WER, to the standard bi-lingual text output without any external information. \\

\end{abstract}

\section{Introduction}

In the recent years, Multilingualism is becoming more widespread due to globalisation and mass migration among people between countries. This makes communication between different communities more difficult due to the gap in both culture and language. Closing the culture gap is harder than it sounds and making people understand each others' languages brings us one step closer to harmony. Even though people with different native tongues may speak the same language (such as English) their varied accents and flavours of their own native tongue usually gets mixed with it. This not only happens with people within the communities but also while conversing with others conversations. This requires the need of not only a bare-bones translation service but one which can translate code-switched speech into monolingual text/speech. 
\\
We have Automatic Speech Recognition (ASR) software to handle translation but Code-Switching speech creates a huge challenge for ASR both in terms of Speech and language models. Mixing of linguistic structures of two languages often leads to an increased loss in the structure of code-mixed sentences which often creates problems in extracting text from speech and language models. With the introduction of Deep Neural Networks, DNNs are now capable of acoustic modelling and have a great capacity of data and language sharing across various languages. 

In our study, we mainly focus on conversations between a Code Switched(CS) and a monolingual speaker to help build a system which helps monolingual people understand the context of the situation. We construct an end-to-end model architecture which mainly focuses on extracting the code-mixed text from Speech using the ASR and then thereby generate the monolingual text from the code-mixed text. Often, the transition from one language to another is difficult to predict, since the position of language switching needs enough supervised data to be able to classify them. Hence, we propose the following methods for overcoming these problems, first, a Deep ASR module along with BERT, and ASR with a Neural Machine Translation model. Therefore, we take input as a Code-switched speech and output a Monolingual text from the language which is more evident.

\section{Related Work}
A lot of effort has gone into researching the process of converting code-switched speech to monolingual text as communication is the key to open all doors and bridge all gaps. Several works have constructed a CS ASR on many different language pairs.
An ASR pipelined model usually contains multiple modules which perform different tasks and have different objectives, and therefore need to trained separately. The usage of end-to-end ASR for code-switching is not new and have been found to perform better than traditional pipelined methods. There are a multitude of architectures that have been used for CS ASR. A few notable ones that we referred to are stated below.\\
The connectionist temporal classification loss function based model by Graves et al. predicted the underlying phonemes without any prior alignment information. \cite{Graves12-ST}
A few proposed architectures have also used CNNs within the end-to-end ASR \cite{ZhangCJ16}\cite{DBLP:journals/corr/ZhangPBZLBC17}\cite{Collbert-W2L}.
Attention based encoder-decoder models have also been used for end-to-end speech recognition \cite{ChorowskiBCB14}\cite{DBLP:journals/corr/ChorowskiBSCB15}\cite{DBLP:journals/corr/BahdanauCSBB15}  along with a few models that use RNN Transducers in their underlying architectures \cite{Graves14}.
For most systems, the training data required is huge in the range of hundreds to thousands of hours. Most of the data available is in monolingual and mostly in English. We require ways to handle not just CS languages but also generate code switched speech or text to augment existing CS Data.
The main issues with CS Translation is not only translation of CS language to an understandable monolingual text but also the detection of multiple languages within the utterance and unpredictable switching positions that can convey different contexts and meanings when translated. 

\section{Proposed Approach}
We propose different methods that we used to first convert CS Speech to CS Text and then CS Text to Monolingual text. We also construct an end-to-end system as part of the approach using NMT model and BERT. Due to a deficiency in code-switched data we try to generate our own CS text via Generative Adversarial Networks, and use the Text-to-Speech API to generate the speech corpus.

\subsection{Speech Recognition Module}
Our Automatic Speech Recognition module is an Attention-based encoder decoder model used to transcribe the acoustic model, using stacked Bidirectional LSTM layers, which takes as input the 1D sequence of speech data generated using Mel-Spectrogram, with 25ms window of signal, that is log scaled and the extracted Short-time Fourier spectrum is passed as a sequence, along with \(h_{t-1}\)  to generate \(h_t\) . 
The Bi-LSTMs are particularly used here to capture the pronunciation of word and how it depends on phonemes of the word in bi-directional context \cite{sivasankaran2018phone}.
The Encoder mainly focuses on transforming the input frames \(x_{1,...,T}\)  to hidden encoder states \(h_{1,..,T}\) and since, as part of CS text, we're concerned with the positions of the usage of words in CS text when converted to monolingual text, we use the attention mechanism to determine the position of the translated work in decoder, based on a context vector generated by capturing the relevance using a dot product between the hidden encoder representations and the \(i^{th}\) decoder state \(s_i\).  
\begin{equation}
\boldsymbol{s}_{i}=\phi\left(\boldsymbol{s}_{i-1}, \boldsymbol{y}_{i-1}, \boldsymbol{C}_{i}\right)
\end{equation}\begin{equation}
C_{i}=\sum_{j=1}^{T} \alpha_{i, j} \cdot h_{j}
\end{equation}
Where, the \(\alpha_{i,j} = Softmax(e_{i,j})\) is the weighted vector representation between softmax output and the encoder output. Thereby, calculating the conditional probability of occurrence of \(y_i\) is conditioned on the \(y_{1,..,i-1}\)  and \(C_i\) .

\begin{figure}[ht]
\includegraphics[width=0.5\textwidth]{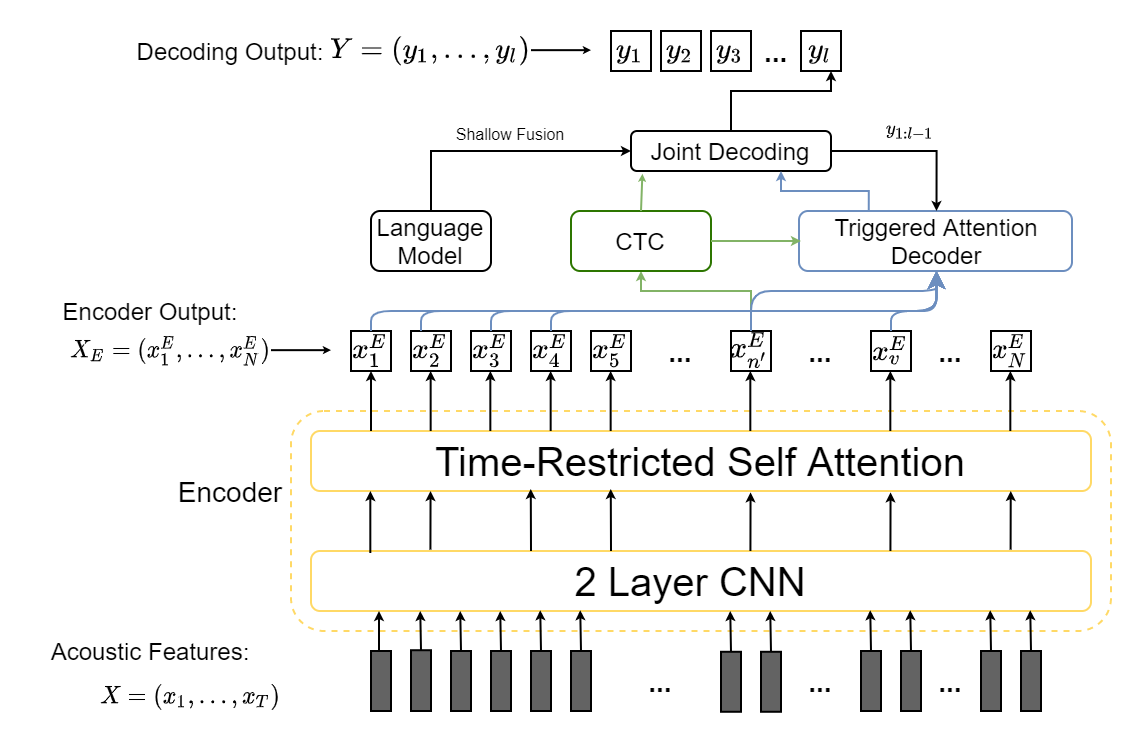}
\caption{Speech Recognition module using hybrid CTC-Transformer model}
\end{figure}
We also worked on using an alternate approach that uses an encoder-decoder network along with CTC (Connectionist  Temporal  Classification) to derive an alignment for our CS text, using which we use CTC output for computing false alignment, and this information is used to truncate the output of encoder state sequence for the attention decoder model accordingly \cite{9054476}. The encoder is a layered network of 2-layer CNN passed through the Transformer encoder as mentioned in the previous method, and then the decoder is now also from Transformer architecture. The triggered attention decoder now allows frame-synchronous decoding, and provides a joint scoring using CTC and attention model for predicting the text from decoder.

\subsection{Code Switched Speech to Monolingual text}
The code switched speech is passed into our neural-ASR architecture which produces the Code-switched text as output. We followed the two models for generating output from our ASR system. The first ASR model was an Attention-based encoder-decoder model. The primary reason of choosing this model was the variable length input and output sequences, which can be accommodated in this model, and generally, most of the code-switched utterances are different in terms of comparison of input and output lengths. The input to the encoder model was log-scaled values of Mel-spectrogram, passed through a FCN using ReLU, into a double-stacked Bi-LSTM Layers consisting of 256 hidden units in each direction. The context vector generated was generated using the attention-model which produces the conditional output representation at decoder, based on the computed vector similarity using attention, for a given time step, thereby producing efficient output unlike the non-attention based model, which only takes final encoder representation context-vector to perform all decoder operations. 
The decoder takes the characters generated as input to the 128-dimensional embedding layer, passed into an LSTM layer with 256 hidden units, with attention to predict the scores with a MLP. The extracted CS text is then passed through BERT, the transformer model to extract the recovered monolingual text.

\subsection{Code Switched text to Recovered text}
\subsubsection{Recovery using BERT}
Most of the general conditional Language Models(LM) are generally uni-directional models, left-right or right-left models, which in most of the practical cases fails to capture the context of the sentence. BERT being a bi-directional transformer model, we mainly utilize the attention based mechanism, which learns contextual embeddings in both directions. We use the pre-trained BERT model and based on Ghazvininejad et al.\cite{Lewis2019BARTDS} we use the mask based prediction algorithm that mainly masks randomly chosen words and predicts it using the language context. Based on Masked LM using BERT, given a code-switched sentence, we particularly mask the unwanted words from the second language \(<MASK>\), and use BERT to recover these words in terms of monolingual text output \cite{lample2019cross}. This task is an unsupervised task, since we particularly feed data to model without any supervised prediction of number of words outputted from the masked words, so post recovery, the difference in the predicted and masked words are filled by padding it with a \(<PAD> \) token. The hyper-parameters of BERT remain the same as the original paper, we use the publicly available BERT(English) model with 12 layered architecture with 768 hidden units, approx. 110-M parameters. The BERT input was curated as above, and didn't need the usage of traditional \(<SEP>\) and \(<CLS>\) tokens, since we particularly used separate sentences \cite{devlin-etal-2019-bert}.  
\begin{equation}
\text { Attention }(Q, K, V)=\operatorname{softmax}\left(\frac{Q K^{T}}{\sqrt{d_{k}}}\right) V
\end{equation}
Based on "the Scaled dot-product attention" \cite{AIAYN}, where given a query Q and key-value pair K and V of dimensions \(d_k\) and \(d_v\), the attention mechanism computes dot product based on the compatibility function of query and key, and divide it by square root of \(d_k\). The positional encoding in BERT are very useful in utilizing the order of the input sequence, where the PE, is generally a sine or cosine function passed into the encoder and decoder stack at the input, to act as positional parameters since, the transformer does not have any convolution or recurrent operations to understand the positions. The multi-head attention allows our LM to understand the content from different sub-spaces at different sub-positions of the given sentence, thereby generating the output recovered text by predicting the masked text in the given sentence. We use the BERT tokenizer for pre-processing the words, and since this tokenizer implements word-piece tokenization, it splits the words into sub-words for unseen words. Hence, it eliminates the out-of-vocabulary problem from our data.

As part of a supervised approach, we also supply the BERT model with both the CS text and it's monolingual text, and randomly mask word tokens in both these text inputs, and now the model tries to use either the surrounding words in the code switched text or the monolingual translation, to thereby encouraging the model to leverage the monolingual translation for context when there is a deficiency in the context from the code switched text, and also encouraging the model in alignment of sentences for predicting the output text. 
\subsubsection{Recovery using NMT}
Even though the transformer architecture used in BERT is very efficient in terms of predicting the masked text, Transformers were generally built for Machine Translation with shorter sequences of approx. 100-200 words length. The self-attention mechanism has a high computational complexity in terms of quadratic order due to all pairs, and is expensive for an acoustic model because of it's high number of frames. Moreover, the positional encoding induce the positional sense in transformers, without which the transformers don't have a sense of positions of words. Hence, we use a attention based encoder-decoder model for recovery of monolingual text using Neural MT model \cite{luong-etal-2015-effective}. Each word is encoded using fixed size vectors or the word embeddings. We pass the input token embeddings into double-stacked BiLSTM architecture, and after the generation of context vector \(C_i\)  at the attention layer, the encoder representations are passed into decoder architecture, which uses a monolingual word embedding with a single layered LSTM with 256 units, to calculate score using MLP of the output monolingual words generated. 
\begin{figure}[ht]
\includegraphics[width=0.5\textwidth]{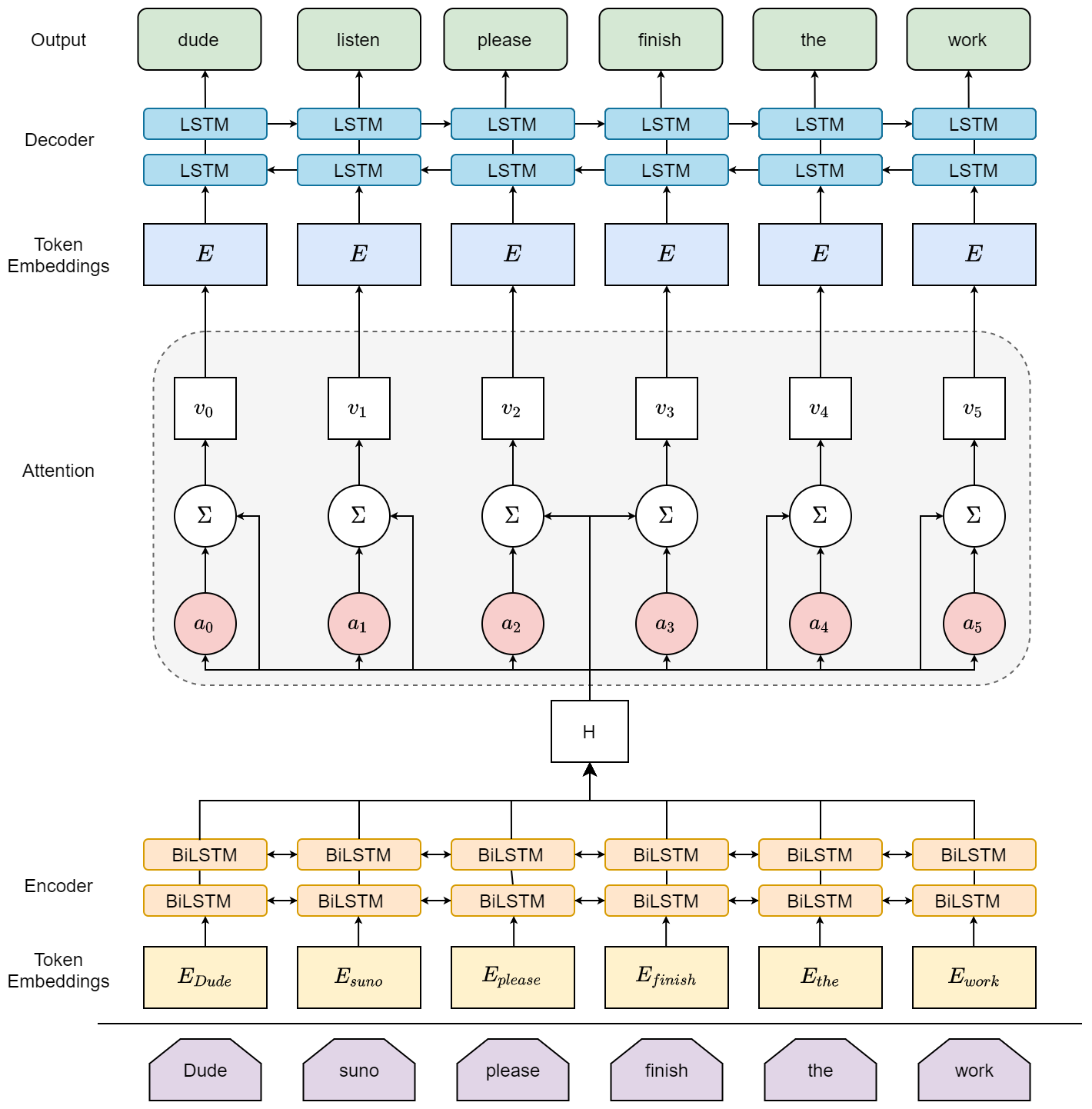}
\caption{End-to-End cascaded model using Encoder-Decoder NMT}
\end{figure}
\subsection{Code Switching Corpus}
We can generate code-switched corpus using two different possibilities, Natural generation and Synthetic generation. Most often, the type of text generated is mostly of types based on the proficiency of the speaker, because of which, it can be deficient-language code switching or efficient language code switching. The deficient case, in most general examples is that case, when given a sentence, only 1-2 words are code switched at different utterances in the sentence, and the latter is the case, where further phrases used in the sentence are code-switched. Hence, we need to keep in mind the different possibilities and generate the corpus similarly. We also know that, the CS text often doesn't have a huge corpus to learn from, and hence, the need to artificially generate the data is important \cite{socialnlp}. The text generated was split into training and test set, and was synthesised using Google Text-to-Speech API. 
\subsubsection{Natural Generation of CS text}
This particular method of generation of corpus is time-consuming, as part of our paper, we do not generate data samples using natural generation. But, for the generation of the dataset, we need a bilingual speaker proficient in both languages, Hindi and English in our case, and selecting 500-1000 pairs of sentences from BTEC and Spoken Language Database, we generate code-mixed sentences by manually translating the sentences, and then, sampling the speech at a frequency of 8 Hz, 16Hz and 32Hz, and determine which particularly produces the best results. The usage of Librosa library \cite{librosa} and extracting STFT from log-scaled Mel Spectrogram at 25ms and 50ms frame-length and passing through the ASR module, would generate results that would be benchmarked against the Synthetic Generation of CS text. 

\subsubsection{Synthetic Generation of CS text}
There are a few ways of generating CS text, one way is to generate the monolingual translations of text in both the primary languages. We then randomly sample the sentences, and based on the dependency parsing, and the trained corpus, replace the most common code-switched words used, and generate samples. Then using the Fuzzy Matching score, for the sentence samples, pick the top-k samples for training corpus. We can also generate multilingual embeddings for the given language along with English, by projecting the word-vectors in the same n-dimensional space, thereby increasing the context of words closely related to each other to be linked by same cluster in the embedding space \cite{pratapa2018language}.

One other way of the artificial way of generation of Code switched sentences is to use Variational Autoencoders (VAEs) \cite{vae}. We design a VAE for code-switched text, that is trained on tagged code-switched sentences, should be able to generate new sentences from the same corpus. We generate the training samples as a pair \(P=(w_i,l_i)\), where \(w_i\)  is the word from the CS text at \(i^{th}\)  position and \(l_i\)  is the language it belongs to. We define the Inference and Generative model as follows:
\begin{equation}
\textbf{Inference model: }q_{\phi}(Z | \boldsymbol{W}, \boldsymbol{Y})
\end{equation}
Given the collection of k code-switched texts, represented by $S^{(m)}=\left(\boldsymbol{w}^{(m)}, \boldsymbol{y}^{(m)}\right): m=1, \ldots, k$ we train the model by maximising ELBO, the evidence lower bound. We use KL divergence based cost annealing used in Bowman et al. to train the system.\cite{bowman}
\begin{equation}
\textbf{Generative model : }p_{\theta}(\boldsymbol{W}, \boldsymbol{Y} | Z)
\end{equation}
The Bayesian rule is applied to the generative model, \(p_\theta (W,Y|Z)\) and then we extract the log \(p_\theta (z)\)  from the expectation term, and after transferring it to KL-Divergence, we optimize the loss function obtained as in the paper, Wang et al. \cite{vae}
\section{Results}

We compared the different methods of end-to-end monolingual text extraction using ASR, using the discussed models. We calculate the Word Error Rate(WER) for following architectures using the artificially generated corpus using the Variational encoders, and we also use the Hindi-English code-mixed dataset scraped from Twitter by Kushagra et al.\cite{Kushagra} We also used the Hindi-English dataset by Vinay et al, both consisting of various social media extracted bilingual sentences along with their translations in English, thereby creating a total of 6500 training examples, and a test-validation split of 500 sentences each. \cite{vinaysingh}

We also generated an artificial corpus, using the encoder-decoder architecture using the VAE, to generate sentences that are code-mixed, which worked well with BERT model since, the sentences produced have a BLEU score of 0.7-0.8, since the model generated sentences modelled based on the training set, along with similar usage of word instances. The rest of the sentences generated using the random sampling of words and replacing with the translation, thereby maintaining a high similarity scores, performed the best in both cases of BERT and NMT, since these sentences had a direct link with the sentences, and context was evident in the sentences, due to lesser masked words in the code-mixed dataset. 

The Artificial keyword in the below table, is mainly referring to the Artificially generated sentences using random switching of words with translations and VAE generated sentences translated using the text2speech API using the methods mentioned in 3.4.1. We also benchmarked the test sets based on different attention-layers, dot-attention being \(s_t^{T}h_i\) , and matrix-attention being the \(s_t^{T}W_ah_i\)  where \(W_a\)  being a trainable weight matrix in the attention layer during training, and \(s_t\)  being the decoder representation at time t, and \(h_i\)  being the encoder representation. 

\begin{table}[th]
\caption{Overall results and performance of ASR+LM on Test data} \label{tab:long}
\begin{tabular}{|l|l|l|}
\hline
\textbf{Model- CS Speech to Mono}         & \textbf{WER(in \%)} & \textbf{PPL} \\ \hline
Artificial+BERT        & \textbf{26.14}               & \textbf{34.219 }           \\ \hline
Artificial+NMT(dot)    & 30.78               & 36.671            \\ \hline
Artificial+NMT(matrix) & 29.4                  & 39.89            \\ \hline
ASR+BERT               & 39.45                  & 51.283            \\ \hline
ASR+NMT(dot)           & 38.22                  & 47.382            \\ \hline
\end{tabular}
\end{table}

\begin{table}[th]
\caption{Results of Language Model(LM) on Test set} 
\begin{tabular}{|l|l|l|}
\hline
\textbf{Model- CS Text to Mono}         & \textbf{WER(in \%)}  \\ \hline
Seq-to-Seq       & 27.43                         \\ \hline
Seq-to-Seq(Attention)   & 19.88                   \\ \hline
Transformer(Vanilla) & 21.67                       \\ \hline
BERT              & 19.45                       \\ \hline
RoBERTa    & 18.32 \\ \hline
% ASR+NMT(dot)           & 38.22                  & 47.382            \\ \hline
\end{tabular}
\end{table}

\section{Conclusion}
We compared the different methods of end-to-end monolingual text extraction using ASR, using Bidirectional Transformers(BERT) and Neural MT model. The results evidently reveal that, error in the extraction of text from code-switched speech using ASR makes the task difficult for BERT, during monolingual extraction. Overall, the model of BERT performed better than Neural Machine Translation in cases of direct code switched text to monolingual text, but when cascaded with the ASR module, the NMT performed better than the ASR+BERT model because the increased number of masked words in the sentences due to the increased error rate in prediction using the ASR than the cleaner CS text available directly. 

In the future, work needs to be done on building a better speech recognition model which works better even for longer sequences, thereby keeping the computational requirements minimum, and also a reduced Word Error Rate for the results on the bench marked tasks.

\bibliographystyle{IEEEtran}

\bibliography{template}

\begin{thebibliography}{10}
\providecommand{\url}[1]{#1}
\csname url@samestyle\endcsname
\providecommand{\newblock}{\relax}
\providecommand{\bibinfo}[2]{#2}
\providecommand{\BIBentrySTDinterwordspacing}{\spaceskip=0pt\relax}
\providecommand{\BIBentryALTinterwordstretchfactor}{4}
\providecommand{\BIBentryALTinterwordspacing}{\spaceskip=\fontdimen2\font plus
\BIBentryALTinterwordstretchfactor\fontdimen3\font minus
  \fontdimen4\font\relax}
\providecommand{\BIBforeignlanguage}[2]{{%
\expandafter\ifx\csname l@#1\endcsname\relax
\typeout{** WARNING: IEEEtran.bst: No hyphenation pattern has been}%
\typeout{** loaded for the language `#1'. Using the pattern for}%
\typeout{** the default language instead.}%
\else
\language=\csname l@#1\endcsname
\fi
#2}}
\providecommand{\BIBdecl}{\relax}
\BIBdecl

\bibitem{Graves12-ST}
A.~Graves, ``Sequence transduction with recurrent neural networks,''
  \emph{arXiv preprint arXiv:1211.3711}, 2012.

\bibitem{ZhangCJ16}
\BIBentryALTinterwordspacing
Y.~Zhang, W.~Chan, and N.~Jaitly, ``Very deep convolutional networks for
  end-to-end speech recognition,'' \emph{CoRR}, vol. abs/1610.03022, 2016.
  [Online]. Available: \url{http://arxiv.org/abs/1610.03022}
\BIBentrySTDinterwordspacing

\bibitem{DBLP:journals/corr/ZhangPBZLBC17}
\BIBentryALTinterwordspacing
Y.~Zhang, M.~Pezeshki, P.~Brakel, S.~Zhang, C.~Laurent, Y.~Bengio, and A.~C.
  Courville, ``Towards end-to-end speech recognition with deep convolutional
  neural networks,'' \emph{CoRR}, vol. abs/1701.02720, 2017. [Online].
  Available: \url{http://arxiv.org/abs/1701.02720}
\BIBentrySTDinterwordspacing

\bibitem{Collbert-W2L}
C.~P. R.~Collobert and G.~Synnaeve, ``Wav2letter: an end-to-end convnet-based
  speech recognition system,'' \emph{CoRR}, vol. vol. abs/1609.03193, 2016.

\bibitem{ChorowskiBCB14}
\BIBentryALTinterwordspacing
J.~Chorowski, D.~Bahdanau, K.~Cho, and Y.~Bengio, ``End-to-end continuous
  speech recognition using attention-based recurrent {NN:} first results,''
  \emph{CoRR}, vol. abs/1412.1602, 2014. [Online]. Available:
  \url{http://arxiv.org/abs/1412.1602}
\BIBentrySTDinterwordspacing

\bibitem{DBLP:journals/corr/ChorowskiBSCB15}
\BIBentryALTinterwordspacing
J.~Chorowski, D.~Bahdanau, D.~Serdyuk, K.~Cho, and Y.~Bengio, ``Attention-based
  models for speech recognition,'' \emph{CoRR}, vol. abs/1506.07503, 2015.
  [Online]. Available: \url{http://arxiv.org/abs/1506.07503}
\BIBentrySTDinterwordspacing

\bibitem{DBLP:journals/corr/BahdanauCSBB15}
\BIBentryALTinterwordspacing
D.~Bahdanau, J.~Chorowski, D.~Serdyuk, P.~Brakel, and Y.~Bengio, ``End-to-end
  attention-based large vocabulary speech recognition,'' \emph{CoRR}, vol.
  abs/1508.04395, 2015. [Online]. Available:
  \url{http://arxiv.org/abs/1508.04395}
\BIBentrySTDinterwordspacing

\bibitem{Graves14}
N.~J. Alex~Graves, ``Towards end-toend speech recognition with recurrent neural
  networks,'' \emph{Proceedings of the 31st International Conference in Machine
  Learning (ICML)}, pp. 1764--1772, 2014.

\bibitem{sivasankaran2018phone}
\BIBentryALTinterwordspacing
S.~Sivasankaran, B.~M.~L. Srivastava, S.~Sitaram, K.~Bali, and M.~Choudhury,
  ``Phone merging for code-switched speech recognition,'' in \emph{Workshop on
  Computational Approaches to Linguistic Code Switching, 2018}, July 2018.
  [Online]. Available:
  \url{https://www.microsoft.com/en-us/research/publication/phone-merging-code-switched-speech-recognition/}
\BIBentrySTDinterwordspacing

\bibitem{9054476}
N.~{Moritz}, T.~{Hori}, and J.~{Le}, ``Streaming automatic speech recognition
  with the transformer model,'' in \emph{ICASSP 2020 - 2020 IEEE International
  Conference on Acoustics, Speech and Signal Processing (ICASSP)}, 2020, pp.
  6074--6078.

\bibitem{Lewis2019BARTDS}
M.~Lewis, Y.~Liu, N.~Goyal, M.~Ghazvininejad, A.~Mohamed, O.~Levy, V.~Stoyanov,
  and L.~Zettlemoyer, ``Bart: Denoising sequence-to-sequence pre-training for
  natural language generation, translation, and comprehension,'' \emph{ArXiv},
  vol. abs/1910.13461, 2019.

\bibitem{lample2019cross}
G.~Lample and A.~Conneau, ``Cross-lingual language model pretraining,''
  \emph{Advances in Neural Information Processing Systems (NeurIPS)}, 2019.

\bibitem{devlin-etal-2019-bert}
\BIBentryALTinterwordspacing
J.~Devlin, M.-W. Chang, K.~Lee, and K.~Toutanova, ``{BERT}: Pre-training of
  deep bidirectional transformers for language understanding,'' in
  \emph{Proceedings of the 2019 Conference of the North {A}merican Chapter of
  the Association for Computational Linguistics: Human Language Technologies,
  Volume 1 (Long and Short Papers)}.\hskip 1em plus 0.5em minus 0.4em\relax
  Minneapolis, Minnesota: Association for Computational Linguistics, Jun. 2019.
  [Online]. Available: \url{https://www.aclweb.org/anthology/N19-1423}
\BIBentrySTDinterwordspacing

\bibitem{AIAYN}
\BIBentryALTinterwordspacing
A.~Vaswani, N.~Shazeer, N.~Parmar, J.~Uszkoreit, L.~Jones, A.~N. Gomez,
  L.~Kaiser, and I.~Polosukhin, ``Attention is all you need,'' \emph{CoRR},
  vol. abs/1706.03762, 2017. [Online]. Available:
  \url{http://arxiv.org/abs/1706.03762}
\BIBentrySTDinterwordspacing

\bibitem{luong-etal-2015-effective}
\BIBentryALTinterwordspacing
T.~Luong, H.~Pham, and C.~D. Manning, ``Effective approaches to attention-based
  neural machine translation,'' in \emph{Proceedings of the 2015 Conference on
  Empirical Methods in Natural Language Processing}.\hskip 1em plus 0.5em minus
  0.4em\relax Lisbon, Portugal: Association for Computational Linguistics, Sep.
  2015, pp. 1412--1421. [Online]. Available:
  \url{https://www.aclweb.org/anthology/D15-1166}
\BIBentrySTDinterwordspacing

\bibitem{socialnlp}
M.~R. Ahan, R.~Honnesh, and M.~Ayush, ``Social network analysis using data
  segmentation and neural networks,'' \emph{” International Research Journal
  of Engineering and Technology (IRJET)}, 2018.

\bibitem{librosa}
B.~McFee, C.~Raffel, D.~Liang, D.~Ellis, M.~Mcvicar, E.~Battenberg, and
  O.~Nieto, ``librosa: Audio and music signal analysis in python,'' in
  \emph{librosa}, 01 2015, pp. 18--24.

\bibitem{pratapa2018language}
\BIBentryALTinterwordspacing
A.~Pratapa, G.~Bhat, M.~Choudhury, S.~Sitaram, S.~Dandapat, and K.~Bali,
  ``Language modeling for code-mixing: The role of linguistic theory based
  synthetic data,'' in \emph{ACM}.\hskip 1em plus 0.5em minus 0.4em\relax ACL,
  July 2018. [Online]. Available:
  \url{https://www.microsoft.com/en-us/research/publication/language-modeling-code-mixing-role-linguistic-theory-based-synthetic-data/}
\BIBentrySTDinterwordspacing

\bibitem{vae}
\BIBentryALTinterwordspacing
P.~Li, W.~Lam, L.~Bing, and Z.~Wang, ``Deep recurrent generative decoder for
  abstractive text summarization,'' \emph{CoRR}, vol. abs/1708.00625, 2017.
  [Online]. Available: \url{http://arxiv.org/abs/1708.00625}
\BIBentrySTDinterwordspacing

\bibitem{bowman}
\BIBentryALTinterwordspacing
S.~R. Bowman, L.~Vilnis, O.~Vinyals, A.~Dai, R.~Jozefowicz, and S.~Bengio,
  ``Generating sentences from a continuous space,'' in \emph{Proceedings of The
  20th {SIGNLL} Conference on Computational Natural Language Learning}.\hskip
  1em plus 0.5em minus 0.4em\relax Berlin, Germany: Association for
  Computational Linguistics, Aug. 2016, pp. 10--21. [Online]. Available:
  \url{https://www.aclweb.org/anthology/K16-1002}
\BIBentrySTDinterwordspacing

\bibitem{Kushagra}
K.~Singh, I.~Sen, and P.~Kumaraguru, ``A twitter corpus for hindi english code
  mixed dataset for pos tagging.'' \emph{ArXiv}, 09 2018.

\bibitem{vinaysingh}
\BIBentryALTinterwordspacing
V.~Singh, D.~Vijay, S.~S. Akhtar, and M.~Shrivastava, ``Named entity
  recognition for {H}indi-{E}nglish code-mixed social media text,'' in
  \emph{Proceedings of the Seventh Named Entities Workshop}.\hskip 1em plus
  0.5em minus 0.4em\relax Melbourne, Australia: Association for Computational
  Linguistics, Jul. 2018, pp. 27--35. [Online]. Available:
  \url{https://www.aclweb.org/anthology/W18-2405}
\BIBentrySTDinterwordspacing

\end{thebibliography}

% \begin{thebibliography}{9}
% \bibitem[1]{Davis80-COP}
%   S.\ B.\ Davis and P.\ Mermelstein,
%   ``Comparison of parametric representation for monosyllabic word recognition in continuously spoken sentences,''
%   \textit{IEEE Transactions on Acoustics, Speech and Signal Processing}, vol.~28, no.~4, pp.~357--366, 1980.
% \bibitem[2]{Rabiner89-ATO}
%   L.\ R.\ Rabiner,
%   ``A tutorial on hidden Markov models and selected applications in speech recognition,''
%   \textit{Proceedings of the IEEE}, vol.~77, no.~2, pp.~257-286, 1989.
% \bibitem[3]{Hastie09-TEO}
%   T.\ Hastie, R.\ Tibshirani, and J.\ Friedman,
%   \textit{The Elements of Statistical Learning -- Data Mining, Inference, and Prediction}.
%   New York: Springer, 2009.
% \bibitem[4]{YourName17-XXX}
%   F.\ Lastname1, F.\ Lastname2, and F.\ Lastname3,
%   ``Title of your INTERSPEECH 2020 publication,''
%   in \textit{Interspeech 2020 -- 20\textsuperscript{th} Annual Conference of the International Speech Communication Association, September 15-19, Graz, Austria, Proceedings, Proceedings}, 2020, pp.~100--104.
% \end{thebibliography}

\end{document}